\documentclass{article}
\makeatletter
\renewcommand{\fnum@figure}{Fig. \thefigure}
\makeatother
\usepackage[utf8]{inputenc} % allow utf-8 input
\usepackage[T1]{fontenc}    % use 8-bit T1 fonts
\usepackage{hyperref}       % hyperlinks
\usepackage{url}            % simple URL typesetting
\usepackage{booktabs}       % professional-quality tables
\usepackage{amsfonts}       % blackboard math symbols
\usepackage{nicefrac}       % compact symbols for 1/2, etc.
\usepackage{microtype}      % microtypography
\usepackage{gensymb}

\usepackage[pdftex]{graphicx}

\usepackage[square,numbers]{natbib}

\usepackage[T1]{fontenc}
\usepackage{lmodern}
\usepackage[left=2cm, right=2cm, top=2cm]{geometry}

\title{A New Tensioning Method using Deep Reinforcement Learning for Surgical Pattern Cutting}
\date{\vspace{-10ex}}

\begin{document}
\maketitle

\let\thefootnote\relax\footnotetext{2019 IEEE International Conference on Industrial Technology (ICIT), Melbourne, Australia.}

%The 20th IEEE International Conference on Industrial Technology, IEEE-ICIT 2019, Melbourne, Australia.

\begin{center}
Thanh Nguyen\textsuperscript{1}, Ngoc Duy Nguyen\textsuperscript{1}, Fernando Bello\textsuperscript{2}, Saeid Nahavandi\textsuperscript{1}\\
\textsuperscript{1}Institute for Intelligent Systems Research and Innovation, Deakin University, Australia\\
\textsuperscript{2}Department of Surgery and Cancer, Imperial College London, UK\\
E-mail: thanh.nguyen@deakin.edu.au. Tel: +61 3 52278281.
\end{center}

\begin{abstract}
Surgeons normally need surgical scissors and tissue grippers to cut through a deformable surgical tissue. The cutting accuracy depends on the skills to manipulate these two tools. Such skills are part of basic surgical skills training as in the Fundamentals of Laparoscopic Surgery. The gripper is used to pinch a point on the surgical sheet and pull the tissue to a certain direction to maintain the tension while the scissors cut through a trajectory. As the surgical materials are deformable, it requires a comprehensive tensioning policy to yield appropriate tensioning direction at each step of the cutting process. Automating a tensioning policy for a given cutting trajectory will support not only the human surgeons but also the surgical robots to improve the cutting accuracy and reliability. This paper presents a multiple pinch point approach to modelling an autonomous tensioning planner based on a deep reinforcement learning algorithm. Experiments on a simulator show that the proposed method is superior to existing methods in terms of both performance and robustness.
\end{abstract}

\section{Introduction}
\label{sec:1}
Manipulation of soft tissues is among the problems of interests of many researchers in the surgical robotics field [1-3]. Surgical scissors are an efficient tool that is normally used to cut through soft tissues [4]. For deformable substances, the deformation behaviours are highly nonlinear and thus present challenges for manipulation and precise cutting [5]. Other surgical tools, e.g. robotic grippers [6, 7], are needed to pinch and tension the soft tissues to facilitate the cutting. The tensioning direction and force need to be adjusted adaptively when cutting proceeds through a predefined trajectory [8, 9].

Surgical pattern cutting skill is one of the requirements for surgical residents, as listed in the Fundamentals of Laparoscopic Surgery (FLS) training suite [10] and for robotic surgery, as included in the Fundamental Skills of Robotic Surgery (FSRS) [11-13]. Automation of surgical tasks can be helpful as it mitigates surgeon load and errors, reduces time, trauma and expenses. Different levels of surgical automation have been studied broadly in the literature [14-19].

Specifically, Shamaei et al. [20] introduced a teleoperated architecture that facilitates the cooperation between human surgeon and autonomous robot to execute complicated laparoscopic surgical tasks. Human surgeon can supervise and intervene the slave robot any time during the operation of surgical tasks. Findings from that study showed the reduction of surgical time when having the collaboration between human and robot compared with performance of a human operator alone. Osa et al. [21] introduced a framework to address two problems of surgical automation in robotic surgery, including the online trajectory planning and the dynamic force control. By learning both spatial motion and contact force simultaneously through leveraging demonstrations, the framework is able to plan trajectory and control force in real time. Experiments with cutting soft tissue and tying knots showed the robustness and stability of the framework under dynamic conditions.

Machine learning in general or reinforcement learning (RL) [22, 23] in particular has been involved in a number of studies for automation of surgical tasks [24, 25]. The ability of RL to solve sequential decision-making problems makes it suitable for automating complicated tasks [26]. Recent development of deep learning [27-29] has made RL as a robust tool to deal with high-dimensional problems [30]. Chen et al. [31] combined programming by demonstration and RL for motion control of flexible manipulators in minimally invasive surgical performance. Experiments on tube insertion in 2D space and circle following in 3D space showed the effectiveness of the RL-based model. Recently, Baek et al. [32] proposed the use of probabilistic roadmap and RL for optimal path planning in dynamic environment. The method was able to perform resection automation of cholecystectomy by planning a path that avoids collisions in a laparoscopic surgical robot system.

Notably, Thananjeyan et al. [9] investigated a method to learn the tensioning force using deep reinforcement learning (DRL), namely trust region policy optimization (TRPO) [33] for soft tissue cutting [34]. The tensioning problem is modelled as a Markov decision process where action set includes 1mm movements of the tensioning arm in the 2D space. The proposed method is evaluated using a simulator that models a planar deformable sheet as a rectangular mesh of point masses [35]. The performance of the proposed method and its competing models is measured by computing the symmetric difference between the predefined pattern and actual cut. The performance obtained from the experiments on multilateral surgical pattern cutting in 2D orthotropic gauze is superior to those of conventional models, e.g. fixed tensioning and analytic tensioning. The method automatically learns the tensioning policy by choosing a fixed pinch point during the entire cutting process regardless of the cutting pattern complexity. This approach has a disadvantage when the cutting pattern is complex as it requires multiple pinch points as cutting proceeds.

In this study, we propose a multiple pinch point approach based on DRL to learn the tensioning policy effectively for surgical gauze cutting. Through this paper, we will analyse and highlight the advantages of our approach compared to existing methods, i.e. no-tension, fixed, analytic, and single pinch point DRL [9]. To facilitate the unbiased comparisons, we use the same simulator as in [9] to model the deformable surgical gauze for experiments. The next section describes in detail the deformable sheet simulator. Section 3 presents the proposed multiple pinch point approach to tensioning policy learning using DRL. Experimental results and discussions are presented in Section 4, followed by conclusions and future work in Section 5. 

\section{Deformable Sheet Simulator}
\label{sec:2}

\begin{figure}[!h]
\centering
\includegraphics[width=0.95\linewidth]{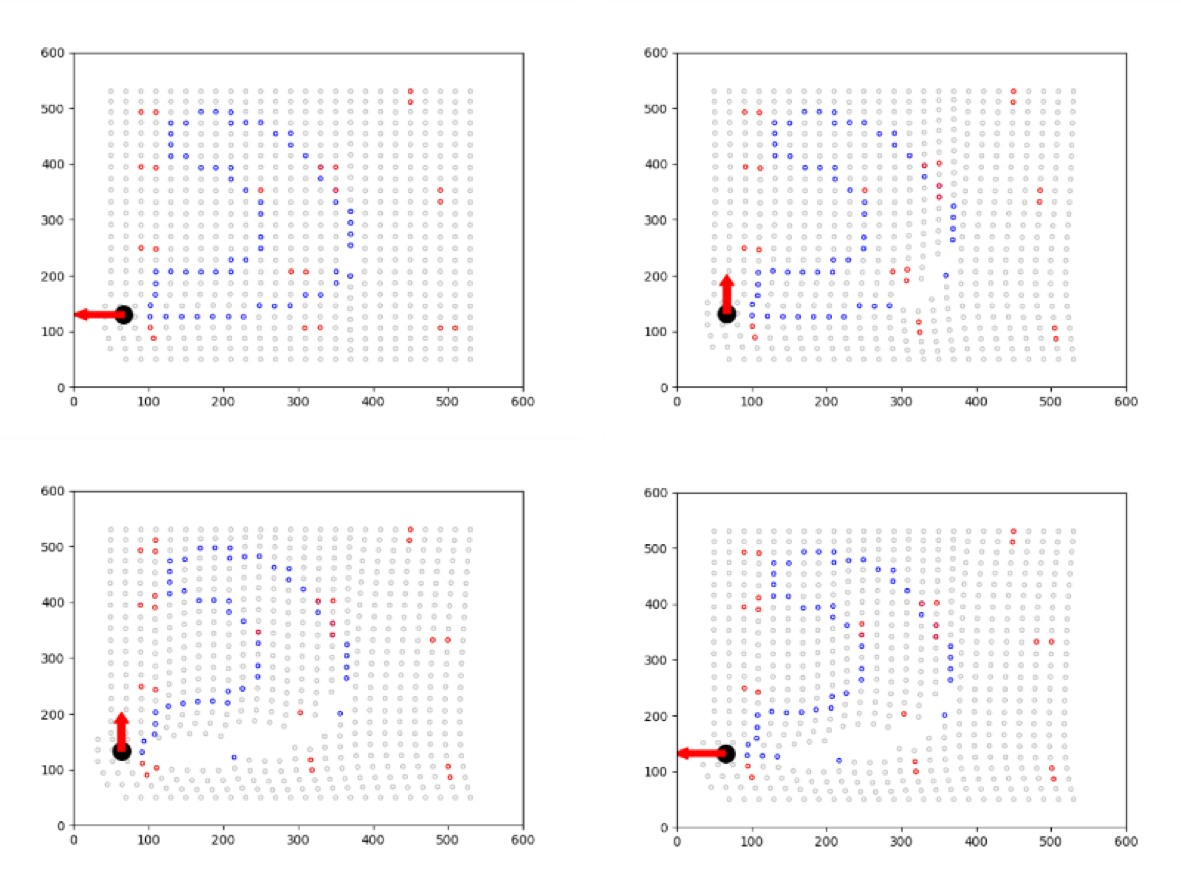}
\caption{Simulation of the deformable sheet for tensioning policy learning.}
\label{fig:1} 
\end{figure}

We use a finite-element simulator to test the proposed algorithm and its competing methods, as with Thananjeyan et al. [9]. The deformable gauze is modelled as a square planar sheet of mesh points whose locations comprise the state of the gauze. When the gauze is tensioned at a pinch point, the mesh points are moved and therefore their coordinates are observed as a new state of the sheet. 

The sheet is initialized with equally spaced mesh points, defined as $\sum$, and locations of these points are denoted as $\sum^{(G)}$ in the global coordinate frame, $G_{x,y,z} \subset R^3$. Thus, the state of the sheet at time point t is defined as $\sum_t^{(G)}$. Movements of the tensioning arm are assumed to be within the $G_{x,y}$ plane, and therefore the initial state of the sheet $\sum_0^{(G)} \in G_{x,y}$. The state is then repeatedly updated through the simulation as: 

$\sum_{t+1}^{(G)} \longleftarrow sim(\sum_t^{(G)})$ \hfill (1)

To simulate the state of the sheet when tensioning, the location of each mesh point $p \in Z$ is updated at each time step [9]: 

$p_{t+1} = \alpha p_t + \delta (p_t - p_{t-1}) - $
$\sum_{(p' \in {neightbors} \wedge   \overline{cut})}\tau (p'_t - p_t) + F_t,$ \hfill (2)

\noindent where $\alpha$ and $\delta$ are time-constant parameters that specify the rate at which the sheet reacts to an applied external force $F_t$, $\tau$ is a spring constant, $\tau(p'_t - p_t)$ is a model that characterizes the interactions between vertices [9]. Cutting is modelled as removing the vertices on the trajectory from the mesh so that these vertices no longer affect their neighbors. 

For a pinch point $s \in Z$, the tensioning problem is specified as constraining location of this pinch point to a position $u \in G_{x,y,z}:T=<s,u>$. The tensioning policy $\pi$ is defined as:

$\pi: \sum_t^{(G)} \longrightarrow \Delta_u$ \hfill (3)

\noindent
where $\delta_u = u_{t+1} - u_t$.

Fig. 1 exhibits the simulation sheets with four different tensioning directions and the reaction of mesh points on the tensioning. Tensioning is required to be adaptive to the deformation of the sheet at each time step of cutting process. In this study, we present a multiple pinch point tensioning approach where the cutting contour is segmented, and each pinch point is used for an individual segment to improve the cutting accuracy. 

\section{Multiple Pinch Points DRL Tensioning Method}
\label{sec:A}

Cutting with scissors can only proceed in the pointing direction of the scissors. In some cases, scissors are required to be rotated all 360$\degree$ degrees to complete a complex contour in a single cut. This may not be possible in cases of robotic surgery where cutting arms are constrained to a rotation limit. Therefore, the cutting contours are normally broken into several segments with each segment is cut with a different starting point, namely a notch point (Fig. 2). 

\begin{figure}[!t]
\centering
\includegraphics[width=0.45\linewidth]{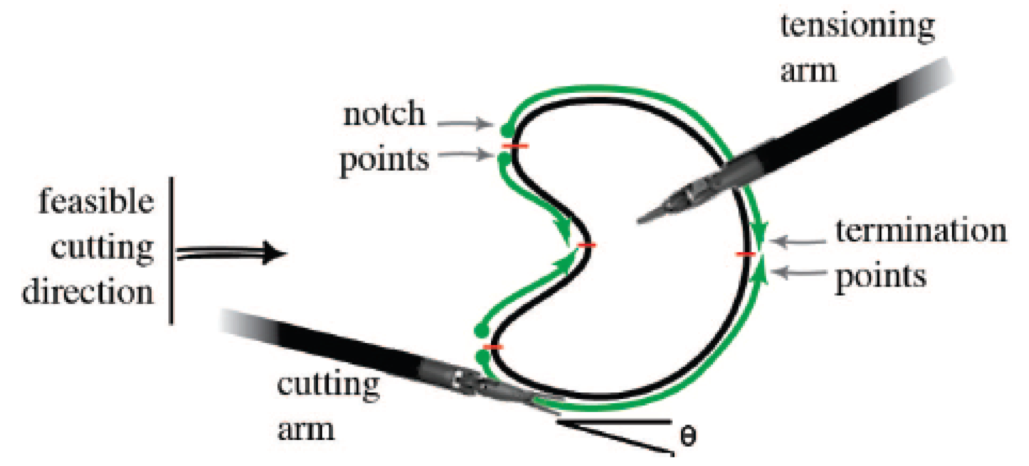}
\caption{The contour is separated into several segments and each segment is cut with a different starting point (notch point) [9].}
\label{fig:2} 
\end{figure}

We first divide the entire cutting trajectory into several segments and choose a suitable pinch point for each segment. The final number of pinch points is therefore equal to the number of segments. We then implement the TRPO algorithm to learn the tensioning policy for each segment based on the corresponding candidate pinch points. Once this step is complete for every segment, we deploy a final aggregate learning step to systematically choose the best pinch point and its corresponding policy for each segment. This is to ensure the whole contour is cut continuously with a smooth transition between segments. This is important because tensioning causes the movement of vertices of the mesh and when each segment is treated separately, the termination point of the previously cut segment need to be matched with a notch point of the next segment. Our approach is diagrammed in detail in Fig. 3.

\begin{figure*}[!t]
\centering
\includegraphics[width=1.0\linewidth]{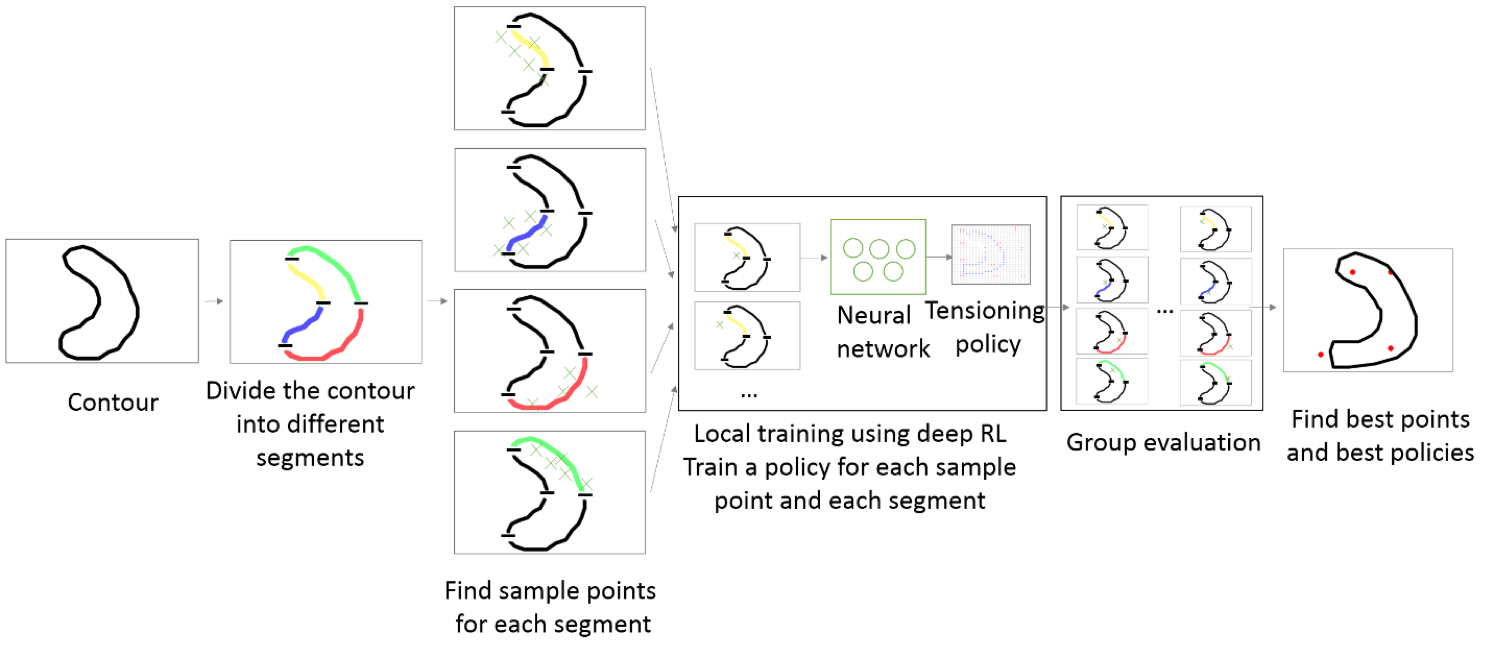}
\caption{Tensioning policy learning by deep RL algorithm with multiple pinch points.}
\label{fig:3} 
\end{figure*}

\subsection{Dynamic Multiple Pinch Point Selection}

Selection of multiple pinch points for multiple segments for tensioning consists of four continuous steps outlined as follows. 

\textit{Step 1: Select candidate pinch points} -- We first divide the contour into N different segments based on local minima and maxima using directional derivatives. We then list all candidate pinch points $\Delta$ satisfying the condition that two arms of the robot (cutting and pinch arms) are not conflicting. These candidate pinch points are grouped into N groups $\Delta_i$ (i=1,2,...,N) (where N is the number of segments) with the following constraint. Given a trajectory $T_i$  of the i--th segment of the contour, a pinch point p in $\Delta$ can serve as a candidate for $T_i$ if:

$min\{ |p-t_k| | \forall  t_k \in T_i\} \leq \delta$ \hfill (4)

\noindent
where $|a-b|$ denotes the Euclidean distance between points a and b, $t_k$ is a point belongs to segment $T_i$, and $\delta$ denotes a distance threshold. In the simulation, we select $\delta$=50 if a contour has more than 2 segments and $\delta$=100 if the contour has no more than 2 segments.

\textit{Step 2: Pruning pinch points} -- To improve the quality of selected pinch points, we prune the redundancy pinch points as follows:

\begin{itemize}

\item	 All direct neighbors of a pinch point are removed.

\item We randomly select 30 pinch points for each segment (10 pinch points if the number of segments is greater than 2). This approach is more robust than the previous study [9] as we only select quality pinch points. It means that we select pinch points that are close to the contour and we prevent multiple pinch points from distributing in the same local area by pruning neighbor pinch points. Therefore, the selected pinch points are uniformly distributed in all areas that are closest to the contour.

\end{itemize}

\textit{Step 3: Local training} -- Train a tensioning policy for each selected pinch point in each segment. This is only a local training, i.e., the policy is trained for cutting only one individual segment. We modify the simulator so that the training is conducted within a designated segment instead of the whole contour in the previous work [9]. This approach significantly reduces the total training time for all candidate pinch points.

\textit{Step 4: Find optimal set of pinch points} -- Find the best order of segments so that the cutting error is minimum. Given a set of segments, there is an optimal order of cutting that minimizes the cutting error. We find the best order of segments by using brute-force search [36]. We go through all segment permutations and perform a cut for each permutation. This is reasonable as the number of segments is normally limited. We assume there is no selected pinch point during the cutting process. The best order will be selected.

Using that order, go through all possible permutations of candidate pinch points to perform the experiment (cutting the entire contour), and select the permutation that provides the best score. We separate the training and evaluate into two steps. This significantly reduces the total time to find the optimal pinch points because it is impossible to train all permutations of selecting pinch points. Therefore, this approach is practical in real-world application. For each pinch point, we have two possible actions: fixed or tensioning. In this study, to reduce the number of configurations, we only apply DRL tensioning for the last pinch point in a permutation while other pinch points are kept with the fixed action. For example, in a set of 4 pinch points (for a contour of 4 segments) of a permutation, $[p_1, p_2, p_3, p_4], p_1, p_2, p_3$ are fixed pinch points, while $p_4$ uses a DRL tensioning policy.

\subsection{Tensioning Policy Learning with DRL}

For unbiased comparisons with existing methods, we propose the use of TRPO algorithm to learn a policy for the tensioning arm. The goal of learning is to minimize the cutting error between the desired contour and the actual cut trajectory. The tensioning problem is modelled as a Markov decision process:

$M = <S,A,\xi,R,T>$, \hfill (5)

\noindent
where $S$ is the state space, $A$ is 1mm movement actions of the tensioning arm in the x and y directions, $\xi$ is unknown dynamics model, $R$ is the reward structure and $T$ is the fixed time horizon. The robot is given zero reward at all time steps except the last step where it receives a reward equivalent to symmetric difference between the marked contour and the actual contour. The reinforcement learning policy $\pi_\theta$ is learned to optimize the expected reward [9]:

$R(\theta) = E_{(s_t,a_t) \sim \pi_\theta}[\sum_{t=0}^T R(s,a)]$ \hfill (6)

We use the TRPO implementation in the rllab framework [37] to optimize $\theta$. The state space is configured as a vector combining the time index of the trajectory $t$, the location of fiducial points selected randomly on the sheet and the displacement vector from the original pinch point $u_t$. The number of fiducial points chosen is 12 for our experiments. This vector is assumed to represent the state of the sheet $\sum_t^{(G)}$ at any time point $t$. A multi-layer perceptron neural network is implemented to map the vector state to a movement action. The network includes two hidden layers with 32 nodes in each layer. 

\section{Simulation Results and Discussions}

\subsection{Cutting Accuracy}

\begin{table}[!t]
\centering
\caption{Experimental results of multiple pinch points DRL versus existing methods.}
\includegraphics[width=0.65\linewidth]{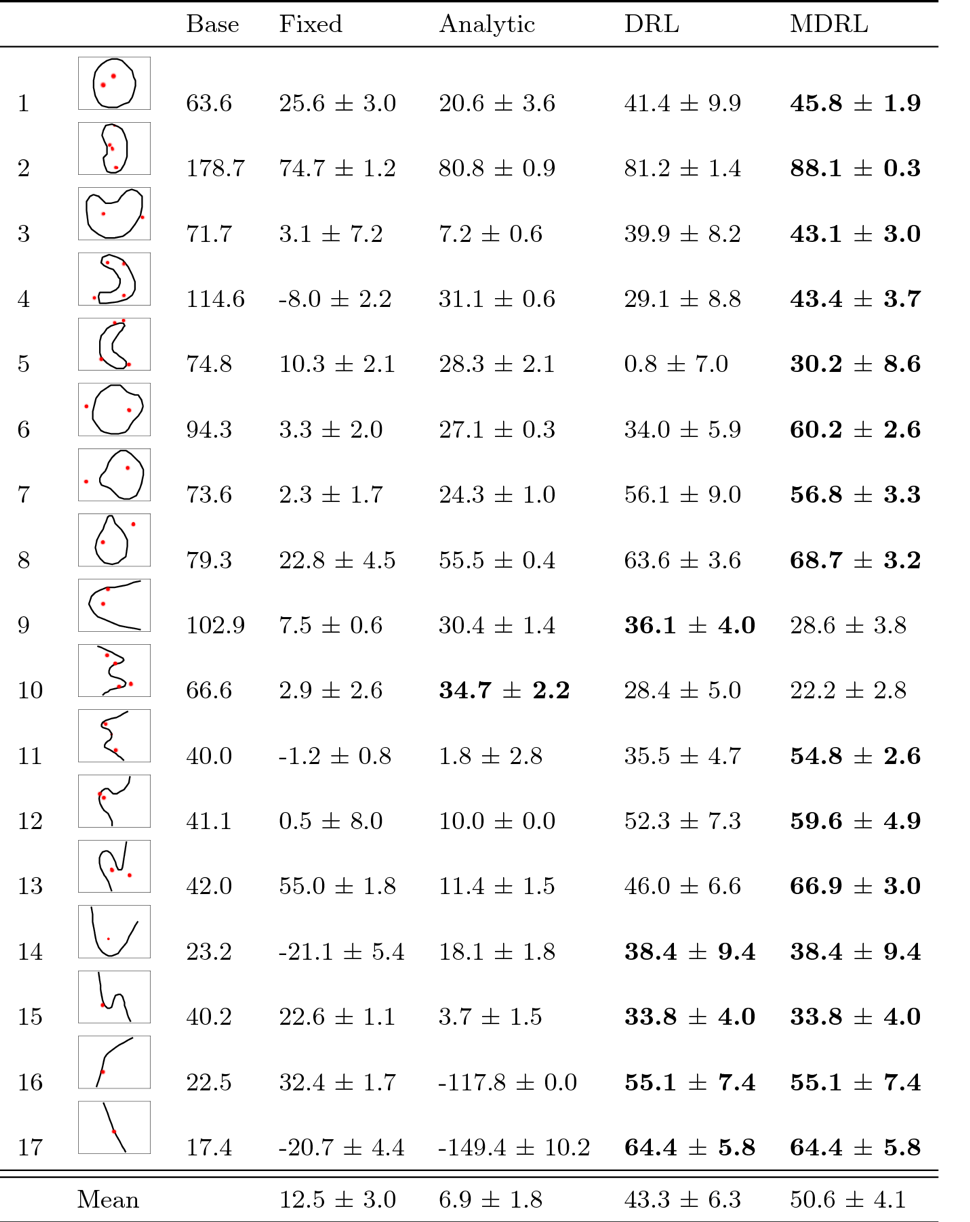}
\label{tab:1} 
\end{table}

We use 17 contour shapes with different complexity levels as in [9] for cutting experiments for ease of comparisons. Cutting accuracy is evaluated based on the symmetric difference between the actual cut contour and the desired contour. We compare our method with four existing methods, including no-tensioning, fixed, analytic and single pinch point DRL [9]. Descriptions of these methods are below:

\begin{itemize}
\item \emph{No-Tensioning}: Cutting proceeds with no assisted tension. The gauze is only suspended with clips at four corners while being cut. 
\item \emph{Fixed Tensioning}: The gauze is pinched at a fixed point when cutting proceeds through the entire contour.
\item \emph{Analytic Tensioning}: The tension is planned based on the direction and magnitude of the error of the cutting tool and the nearest point on the marked contour. 
\item \emph{Single Pinch Point DRL}: A tensioning policy is learned based on DRL uses a single pinch point for the entire contour. This method was proposed in [9].
\item \emph{Multiple Pinch Point DRL (MDRL)}: The contour is broken into several segments and each segment is cut using a tensioning policy with a corresponding pinch point. This method is described in detail in Section 3. The total time to find the optimal policy using MDRL is reasonable and can be controlled by the parameter $\delta$, which is the distance threshold in Eq. (4). Depending on the number of segments, number of selected pinch points, and the value of $\delta$, the whole process takes 6 to 15 hours to find the optimal set of pinch points. Because we separate the training and evaluation into two steps, it is possible to accelerate the process by scheduling them running in parallel.
\end{itemize}

We learn a policy by 20 iterations with a batch size of 500 for each contour shape. After training, each shape is tested 20 times with the learned policy and the results in terms of average accuracy are reported. The simulation results of five competing tensioning methods on 17 contours are presented in Table 1. The red dots on each shape represent the optimal pinch points chosen by our algorithm. Each segment has a corresponding pinch point, which is different to the single pinch point method where the entire contour has only one pinch point. The no-tensioning method is chosen as a baseline where its results are presented in terms of symmetric difference between marked contour and actual cut contour. Results of other methods are reported as the improvement percentage against this baseline method. The proposed MDRL outperforms all existing methods in terms of average accuracy. It achieves the improvement of 50.6\% against the baseline method whilst the single pinch point DRL obtains the improvement of 43.3\%. The standard deviation of the MDRL method at 4.1 is smaller than that of the DRL method at 6.3. This demonstrates that the proposed multiple pinch point method is more stable than the single pinch point DRL method. The last four shapes demonstrate the equivalent performance between single pinch point DRL and multiple pinch point DRL because there is only one segment is created for each of these contours. Therefore, there is only one single pinch point is generated for the entire contour and thus our proposed method is diminished to the single pinch point method. 

Fig. 4 shows the graphical comparisons of four competing methods where MDRL dominates all other methods in terms of average performance. On average, the analytic method is the worst performer while the single pinch point DRL method is the most unstable method as its standard deviation is the greatest among the competing methods. 

\begin{figure}[!t]
\centering
\includegraphics[width=0.45\linewidth]{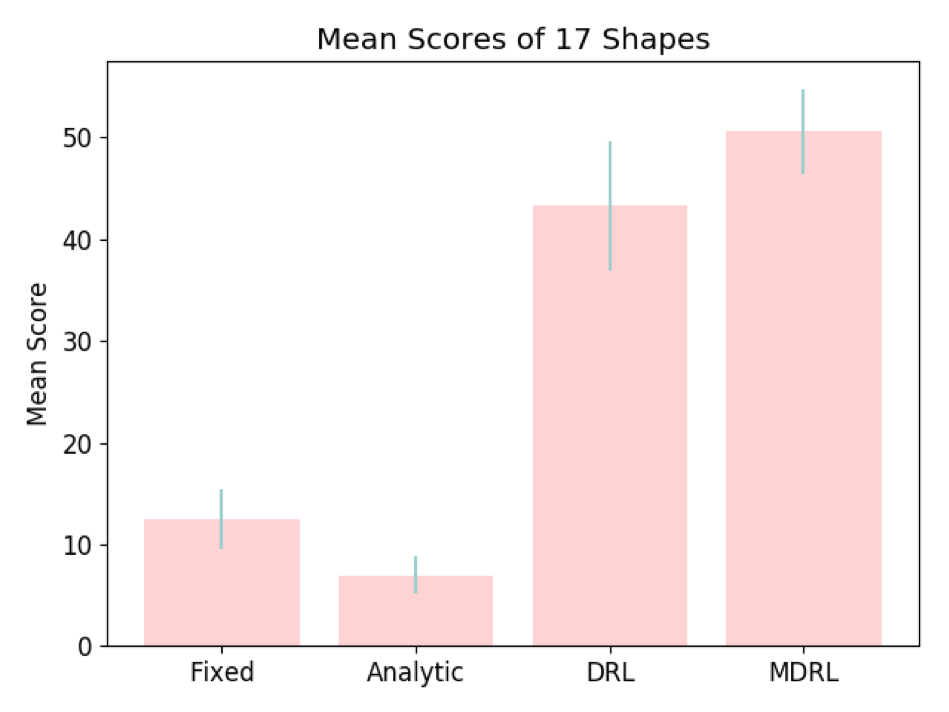}
\caption{Average performance of four competing methods.}
\label{fig:4} 
\end{figure}

\subsection{Robustness Testing}

Learning an autonomous policy based on DRL using a simulation environment would be vulnerable to overfitting. Thananjeyan et al. [9] checked the robustness of the single pinch point DRL tensioning method on different resolutions of the simulated sheet, which were characterized by the number of vertices representing the sheet. The test results on six different resolutions from 400 to 4000 vertices showed that policies on low resolutions can still yield good performance and suggested the use of 625 (25x25) vertices to obtain the greatest cutting accuracy. Therefore, in this study, we use this resolution setting, i.e. 25x25 vertices, to evaluate our algorithm and test the robustness of the competing methods on varying process noise and gravity force.

\begin{figure}[!h]
\centering
\includegraphics[width=0.4\linewidth]{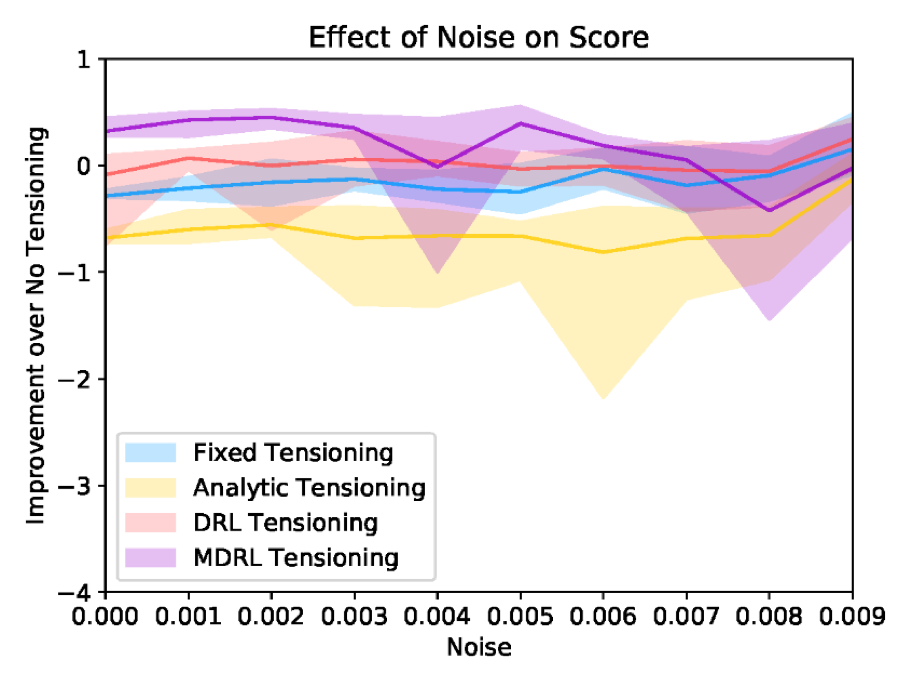}
\caption{Robustness of tensioning methods to the process noise.}
\label{fig:5} 
\end{figure}

Different noise levels are added into the mesh point update formula, i.e. Eq. (2), with the Gaussian noise $N(0,\delta)$ is used for simulation. We run 10 trials for each of 10 values of $\delta$. The effect of noise on performance score of different methods applied on shape 11 is presented in Fig. 5. It is shown that the MDRL is superior to other methods in terms of improvement over the no-tensioning method. This dominance demonstrates the robustness of the MDRL against the process noise injected into the simulation environment. 

\begin{figure}[!t]
\centering
\includegraphics[width=0.4\linewidth]{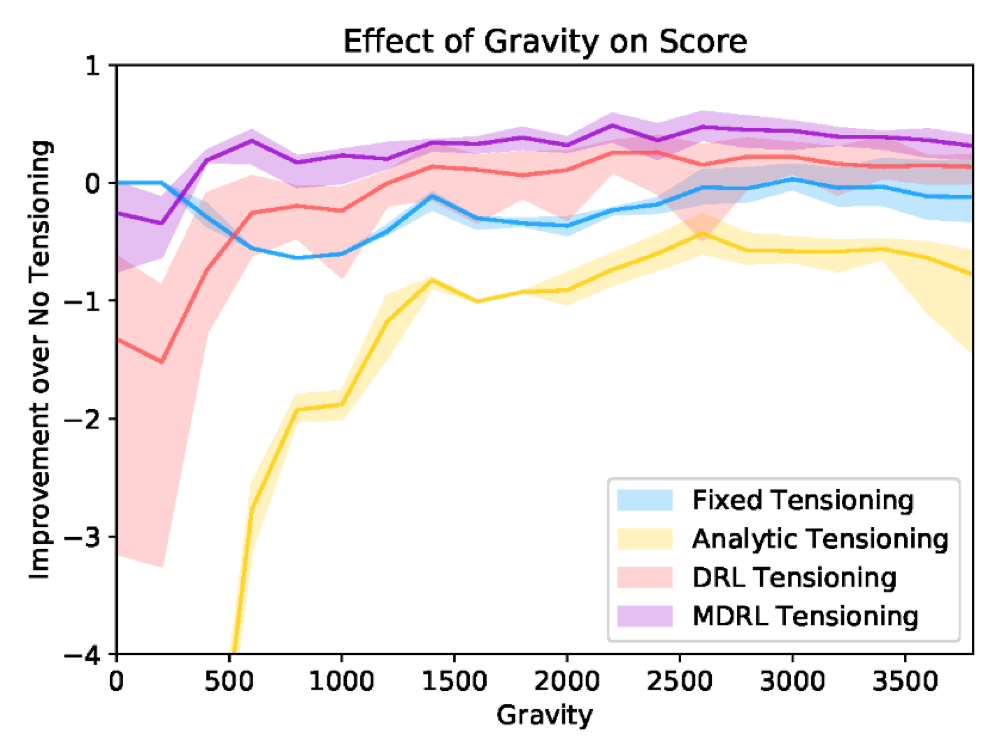}
\caption{Robustness of tensioning algorithms to the gravity force.}
\label{fig:6} 
\end{figure}

We use $f(t)=2500$ to learn tensioning policies for the single pinch point DRL and MDRL methods. All competing methods are then tested on different force magnitudes, ranging from 0 to 3800. The results in terms of improvement over the no-tensioning method using shape 11 are presented in Fig. 6. Clearly, the proposed MDRL method outperforms all other methods as it achieves the best performance over the entire testing range of gravity force. The single pinch point DRL method is the second-best algorithm whilst the analytic tensioning technique is completely dominated by the remaining methods. 

\section{Conclusions}

This paper presents a dynamic multiple pinch point approach to learning effectively a time-varying tensioning policy based on DRL for cutting a deformable surgical tissue. The proposed MDRL algorithm has been tested on a deformable sheet simulator and its performance has been compared to several existing methods. Simulation results demonstrate the significant superiority of our method against its competing methods. Different process noise levels and external gravity forces were added to the simulation environment to test the robustness of the proposed MDRL method. Experimental results show that MDRL is robust to noise and external force and its robustness level is superior to its competing methods. The current method has a disadvantage regarding the number of tensioning directions, which limit at four in this study. A future work would be to increase the number of tensioning directions to improve the cutting accuracy of robot. This would be a necessary natural extension because tensioning in practice would extend to an arbitrary direction depending on the trajectory complexity and the deformation level of the tissue.

\end{document}